\documentclass[runningheads]{llncs}
\usepackage[numbers,sort&compress]{natbib}
\usepackage{amsmath,amssymb,amsfonts}
\usepackage{algorithmic}
\usepackage{graphicx}
\usepackage{textcomp}
\usepackage[svgnames,table]{xcolor}
\def\BibTeX{{\rm B\kern-.05em{\sc i\kern-.025em b}\kern-.08em
    T\kern-.1667em\lower.7ex\hbox{E}\kern-.125emX}}
    
\usepackage{multirow}
\usepackage{makecell}
\usepackage{booktabs}

\renewcommand{\bf}[1]{\cellcolor{black!20}\textbf{#1}}
    
\usepackage{optidef}

\usepackage{pgfplots}
\usepackage{tikz}
\usepgfplotslibrary{statistics}
\pgfplotsset{compat=1.17} 
\usetikzlibrary{calc,positioning}
\usetikzlibrary{shapes.geometric, arrows}
\usetikzlibrary{shapes,shadows,trees,patterns}
\tikzset{
  unit/.style={font=\large, minimum size=0.6cm, draw, fill=white},
  state/.style={
        circle,
        thick,
        draw=black,
        fill=red!50,
        minimum size=6mm,
    },
}
\definecolor{darkred}{rgb}{0.55, 0.0, 0.0}
\definecolor{darkgoldenrod}{rgb}{0.72, 0.53, 0.04}
\definecolor{darkkhaki}{rgb}{0.74, 0.72, 0.42}
\definecolor{navyblue}{rgb}{0.0, 0.0, 0.5}
\definecolor{moonstoneblue}{rgb}{0.45, 0.66, 0.76}
\definecolor{ashgrey}{rgb}{0.7, 0.75, 0.71}
\definecolor{blond}{rgb}{0.98, 0.94, 0.75}
\definecolor{burntumber}{rgb}{0.54, 0.2, 0.14}
\definecolor{darkcyan}{rgb}{0.0, 0.55, 0.55}
\usetikzlibrary{shapes,shadows,trees,patterns}

\usetikzlibrary{arrows.meta,
                matrix,
                positioning}
                
                \usetikzlibrary{matrix,decorations.pathreplacing,calc}
 \pgfkeys{tikz/mymatrixenv/.style={decoration=brace,every left delimiter/.style=    {xshift=3pt},every right delimiter/.style={xshift=-3pt}}}
 \pgfkeys{tikz/mymatrix/.style={matrix of  nodes,left delimiter=[,right delimiter=    {]},inner sep=2pt,column sep=1em,row sep=0.5em,nodes={inner sep=0pt}}}
 
 \pgfkeys{tikz/mymatrixbrace/.style={decorate,thick}}
 
 \newcommand\mymatrixbraceoffsetv{0.2em}

\newcommand*\mymatrixbracetop[4][m]{
\draw[mymatrixbrace] ($(#1.north west)!(#1-1-#2.north west)!(#1.north east)+(0,\mymatrixbraceoffsetv)$)
    -- node[above=2pt] {#4} 
    ($(#1.north west)!(#1-1-#3.north east)!(#1.north east)+(0,\mymatrixbraceoffsetv)$);
}
\newcommand*\mymatrixbracebottom[4][m]{
\draw[mymatrixbrace] ($(#1.south west)!(#1-1-#3.south east)!(#1.south east)-(0,\mymatrixbraceoffsetv)$)
    -- node[below=2pt] {#4} 
    ($(#1.south west)!(#1-1-#2.south west)!(#1.south east)-(0,\mymatrixbraceoffsetv)$);
}

\usepackage{svg}
\usepackage{graphicx}
\graphicspath{{./papers/iscmi_2022_conference_paper/Images/}}
\begin{document}

\title{Exploring the effectiveness of surrogate-assisted evolutionary algorithms on the batch processing problem}

\author{Mohamed Z. Variawa\inst{1}\orcidID{0000-0002-3345-2754} \and 
Terence L. Van Zyl\inst{2}\orcidID{0000-0003-4281-630X} \and 
Matthew Woolway\inst{3}\orcidID{0000-0002-4902-851X}}

\institute{University of Johannesburg, Academy of Computer Science and Software Engineering, Johannesburg, SA,\\
\email{201472992@student.uj.ac.za}
\and
University of Johannesburg, Institute for Intelligent Systems,\\
Johannesburg, SA \\ \email{tvanzyl@gmail.com}
\and
University of Johannesburg, Faculty of Engineering and the Built Environment,\\
Johannesburg, SA \\ \email{matt.woolway@gmail.com}
}

\maketitle

\begin{abstract}
    Real-world optimisation problems typically have objective functions which cannot be expressed analytically. These optimisation problems are evaluated through expensive physical experiments or simulations. Cheap approximations of the objective function can reduce the computational requirements for solving these expensive optimisation problems. These cheap approximations may be machine learning or statistical models and are known as surrogate models. This paper introduces a simulation of a well-known batch processing problem in the literature. Evolutionary algorithms such as Genetic Algorithm (GA), Differential Evolution (DE) are used to find the optimal schedule for the simulation. We then compare the quality of solutions obtained by the surrogate-assisted versions of the algorithms against the baseline algorithms. Surrogate-assistance is achieved through Probablistic Surrogate-Assisted Framework (PSAF). The results highlight the potential for improving baseline evolutionary algorithms through surrogates. For different time horizons, the solutions are evaluated with respect to several quality indicators. It is shown that the PSAF assisted GA (PSAF-GA) and PSAF-assisted DE (PSAF-DE) provided improvement in some time horizons. In others, they either maintained the solutions or showed some deterioration. The results also highlight the need to tune the hyper-parameters used by the surrogate-assisted framework, as the surrogate, in some instances, shows some deterioration over the baseline algorithm.
    \keywords{single-objective optimisation, machine learning, evolutionary algorithms,surrogate models}
\end{abstract}


\section{Introduction}

Single-objective optimisation problems (SOP) are a common occurrence in numerous fields. Most optimisation problems encountered in practice have objective functions that cannot be assessed analytically and call for time-consuming physical experiments or simulations~\cite{YJin2021,perumal2020surrogate, van2021parden}.

Search heuristics called evolutionary algorithms (EA) produce solutions to optimisation and search issues. Evolutionary algorithms employ inheritance, mutation, selection, and crossover methods that are modelled after natural evolution. The computing restrictions of evolutionary algorithms are identical to those of physical experiments or simulations, as often, it may take tens of thousands of fitness assessments to find workable solutions \cite{YJin2021}.

It is possible to use computationally affordable approximations of the goal functions to get around these restrictions while maintaining the quality of the solutions. These low-cost estimates are known as \textit{surrogate models} or \textit{metamodels} (these terms may be used interchangeably) \cite{QChen2022, FJimenez2022,XMa2022, DGuo2022, HHorii2021, JTong2021, HLi2021, stander2020data, stander2020extended, stander2022surrogate}. These stand-ins could be statistical models (like the Gaussian Process) or machine learning models (e.g. artificial neural networks). These surrogates can be trained using historical data or simulation runs that have been carefully chosen.

Previously, \cite{stander2020data} employed surrogate-assisted techniques for the optimisation of design parameters of a chemical process. The authors compared a simulation-only approach to a surrogate-assisted model \cite{stander2020data}. The authors note that the surrogate-assisted achieves a max value faster and is equally able to achieve a better max revenue than the simulation-only model \cite{stander2020data}. Further, the authors demonstrate that surrogate-assisted Genetic Algorithms can scale into increasingly complex systems with parallel and feedback components, with significant speedups and robust results \cite{stander2020extended}.

In this paper, a simulation of the (flowshop) batch-processing problem \cite{MIerapetritou1998} is designed, which accepts a set of instructions indicating which processes should run at which time (herewith referred to as a schedule). The metaheuristic optimisation algorithms Genetic Algorithm (GA) and Differential Evolution (DE) obtain optimal schedules for the batch-processing problem, the baseline algorithms. The objective function used by GA and DE is the simulation of the problem. Then, surrogate-assisted versions of the GA and DE derive optimal schedules for the batch-processing problem. Quality indicators such as Success Rate (SR), Average Evaluations to a Solution (AESR) and Average Generations to a Solution (AGSR) compare the performance of the surrogate-assisted algorithms to the baseline algorithms.

Specifically, the aims of this paper are:
\begin{enumerate}
    \item to show that it is viable to create a simulation to model a well known (flowshop) batch processing problem using the SimPy package,
    \item to illustrate the viability of using the simulation as the objective function of an optimisation problem, and
    \item to investigate the improvements, if any, of using surrogate-assisted evolutionary algorithms as opposed to the sole use of evolutionary algorithms.
\end{enumerate}

\subsection{Single-Objective Optimisation}

The mathematical formulation of a (maximisation) single-objective optimisation problem, is given by:

\begin{maxi}|s|
    {}{f(\textbf{x})}{}{}
     \addConstraint{g_j(\textbf{x})}{\leq 0,}{j=0,\ldots,J}
     \addConstraint{h_k(\textbf{x})}{=0,}{k=0,\ldots,K}
     \addConstraint{\textbf{x}^{L} \leq \textbf{x} \leq \textbf{x}^{U}}{}{}
\end{maxi}\label{eqn:example_mop}

where $f_{\textbf{x}}$ is the objective function, $\textbf{x} \epsilon R^{n}$ is the decision vector, $\textbf{x}^{L}$ and $\textbf{x}^{U}$ are the lower and upper bounds of the decision vector, $\textit{n}$ is the number of decision variables, $g_{j}(\textbf{x})$ are the inequality constraints, and $h_{k}(\textbf{x})$ are the equality constraints, $\textit{J, K}$ are the number of inequality and equality constraints, respectively \cite{YJin2021}. If a solution $\textbf{x}$ satisfies all constraints, it is called a feasible solution, and the solutions that achieve the maximum value are called the optimal solutions \cite{YJin2021}.






\subsection{Probabilistic Surrogate-Assisted Framework (PSAF)}\label{sec:psaf}

The probabilistic surrogate-assisted framework (PSAF) employed in this research was introduced by \cite{JBlank2021} and also forms part of the authors' pysamoo package, which is explored in this research \cite{JBlank2022b}. PSAF may only be applied to single-objective, unconstrained optimisation problems \cite{JBlank2021} (PSAF has been extended for multi-objective, constrained problems in a framework known as GPSAF \cite{JBlank2022a}). The baseline algorithm (i.e. the meta-heuristic that the surrogate will enhance) used in PSAF may be one of many evolutionary algorithms, such as Genetic Algorithm (GA), Particle Swarm Optimisation (PSO), or Covariance matrix adaptation evolution strategy (CMA-ES) \cite{JBlank2021}.

PSAF uses the whole search pattern (i.e. using all the solutions and their offspring) to optimise the surrogate, as opposed to other frameworks which only use the final solution(s). The exploration-exploitation balance is discovered by using the search pattern and accounting for the surrogate's accuracy. PSAF consists of two phases to enable even more adaptable use of the surrogate. The first phase, known as $\alpha$-phase, derives a solution set influenced by the surrogate \cite{JBlank2021}. The second phase, $\beta$-phase, introduces bias by optimising the surrogate for a few iterations \cite{JBlank2021}.


The $\alpha$-phase incorporates a popular concept among evolutionary algorithms known as tournament selection. A population member must win a tournament to participate in the mating process. The quantity of competitors ($\alpha$) balances how greedy the selection process will be. On the one hand, a higher value of $\alpha$ restricts mating to elitist solutions, whereas a lower value lessens the selection pressure \cite{JBlank2021}. The most commonly used tournament mode for genetic algorithms is the binary tournament ($\alpha$=2), which compares a pair of solutions regarding one or multiple metrics. A binary tournament declares the least infeasible solution (i.e. the solution whose objective value is closest to the objective function) as the winner if one or both solutions are infeasible. If both solutions are feasible, the solution with the smaller function value is the winner. In PSAF, tournament selection compares solutions evaluated on the surrogate; this introduces surrogate bias while generating new infill solutions \cite{JBlank2021}. 

  
Although tournament selection effectively incorporates the surrogate's approximation, it is limited by only looking at one iteration into the future. During the $\beta$-phase, the baseline algorithm is run for extra consecutive $\beta$ iterations on the surrogate's approximation, which increases the surrogates' impact. The surrogate's optimum will continuously be reached if $\beta$ is incorrect and will completely discard the baseline algorithm's default infill method. An incorrectly chosen $\beta$ also reduces the infill options' diversity and does not consider the approximation inaccuracy of the surrogate \cite{JBlank2021}. 

The optimisation of the batch-processing problem can be written as, \cite{RSeid2012}:
\begin{maxi}|s|
    {}{\Sigma_{s} price(s^{p})qs(s^{p}, p), \forall p=P, s^{p} \epsilon S^{p}}{}{}
     \addConstraint{tu(s_{in,j^{'}},p)}{\geq tp(s_{inj},p),}{\forall j \epsilon J, p \epsilon P, s_{inj} \epsilon S^{sp}_{inJ^{'}}, s_{inj^{'}} \epsilon S^{sc}_{inJ}}
     \addConstraint{q_{s}(s,p)}{\leq QS^{U}}{\forall s \epsilon S, p \epsilon P}
\end{maxi}\label{eqn:batch_processing_sop}

\subsection{Batch processing problem}\label{sec:case_study}

\tikzstyle{block} = [rectangle, draw, fill=darkgoldenrod!30,
text width=6em, text centered, minimum height=2em, drop shadow]
\tikzstyle{line} = [draw, -latex']
\tikzstyle{cloud} = [draw, circle,fill=gray!60, node distance=3cm,
minimum height=0.5em, drop shadow]
\tikzstyle{circ} = [draw, circle, fill=darkred!50 ,text width=4em, text centered, drop shadow]
\tikzstyle{blockB} = [rectangle, draw, fill=white!60,
text width=8em, text centered, minimum height=4em, drop shadow]

\begin{figure*}[htb!]
    \centering
    \scalebox{0.61}{
        \begin{tikzpicture}[node distance =3.0 cm,scale = 1]
            \node [circ,scale=1] (s1) {\Large$\mathbf{s_1}$};
            \node[block, right of=s1] (mixing) {\textbf{Mixing}};
            \node [circ, right of=mixing,scale=1] (s2) {\Large$\mathbf{s_2}$};
            \node[block, right of=s2] (reaction) {\textbf{Reaction}};
            \node [circ, right of=reaction,scale=1] (s3) {\Large$\mathbf{s_3}$};
            \node[block, right of=s3] (purify) {\textbf{Purification}};
            \node [circ, right of=purify,scale=1] (s4) {\Large$\mathbf{s_4}$};
            \path [line,ultra thick] (s1) -> (mixing);
            \path [line,ultra thick] (mixing) -> (s2);
            \path [line,ultra thick] (s2) -> (reaction);
            \path [line,ultra thick] (reaction) -> (s3);
            \path [line,ultra thick] (s3) -> (purify);
            \path [line,ultra thick] (purify) -> (s4);

        \end{tikzpicture}
    }
    \caption{State Task Network diagram.}
    \label{fig:stn}
\end{figure*}
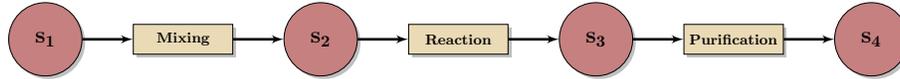

\begin{table}[htbp]
    \caption{System constraints data.}
    \begin{center}
        \begin{tabular}{ccccc}
            \toprule
            \textbf{Unit}  & \textbf{Capacity} & \textbf{Suitability} & \textbf{Time} & \textbf{Price} \\[3pt]
            \toprule
            Unit 1         & 100               & Mixing               & 4.5           & 0              \\[3pt]
            Unit 2         & 75                & Reaction             & 3.0           & 0              \\[3pt]
            Unit 3         & 50                & Purification         & 1.5           & 0              \\[3pt]
            \toprule
            \textbf{State} & \textbf{Storage}  & \textbf{Initial}     &               & \textbf{Price} \\[3pt]
            \toprule
            State 1        & Unlimited         & Unlimited            &               & 0              \\[3pt]
            State 2        & 100               & 0.0                  &               & 0              \\[3pt]
            State 3        & 100               & 0.0                  &               & 0              \\[3pt]
            State 4        & Unlimited         & 0.0                  &               & 0              \\[3pt]
            \bottomrule
        \end{tabular}
        \label{tab:batch_processing}
    \end{center}
\end{table}

This paper's example used as a case study is a batch processing problem described by \cite{MIerapetritou1998, MWoolway2018, MWoolway2019}. The problem studies the production of a single product via three processes: mixing, reaction, and purification. Table \ref{tab:batch_processing} shows the relevant constraints on the system, and the STN in Figure \ref{fig:stn} illustrates the flow of the product through the system \cite{MWoolway2019}. Both fixed and variable processing time variants of this problem have been solved using evolutionary algorithms \cite{MWoolway2018, MWoolway2019, bowditch2019comparative}. The Makespan minimisation problem has also been solved using evolutionary algorithms \cite{TVanZyl2020}. The global solutions (i.e. the optimal schedules for various time horizons) employed in this paper have been reported in the literature \cite{MWoolway2018, MWoolway2019, TVanZyl2020}. The fixed times are shown in Table \ref{tab:batch_processing} while the variable processing times are batch dependent (i.e. the time depends on the amount of product) and not considered in this paper. Table \ref{tab:batch_processing} shows the storage constraints of each storage unit in the system as well as the capacity constraints (i.e. how much material each process can take in). A simulation of the process in the STN diagram \ref{fig:stn} is created and used as the objective function in this research, described further in Section \ref{sec:simulation}. The simulation output is the amount of product produced by a solution for a given time horizon. The solutions take the form of binary instruction vectors which tell the simulation which process to run at which time. The decision variables are binary digits which indicate whether a process should be run ($1$) or not ($0$). The instruction vectors and decision variables are discussed in more detail in Section \ref{sec:simulation}.

\section{Methodology}

\subsection{Setting up the simulation}\label{sec:simulation}

SimPy is a Discrete-Event Simulator which allows for the creation of simulations based on (discrete) real-world processes \cite{NMatloff2008}. A simulation in SimPy is created using the parameters described in Section \ref{sec:case_study} and the State Task Network (STN) diagram shown in Figure \ref{fig:stn}. The units and states in Table \ref{tab:batch_processing} are Containers in the simulation. The three processes (mixing, reaction, and purification) are Python functions which control the contents of the Containers.

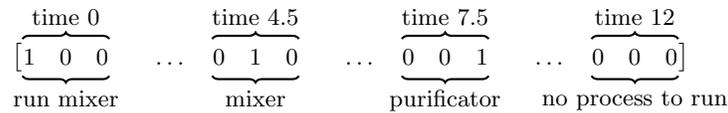
\begin{figure*}[htb!]
\centering
    \begin{tikzpicture}[mymatrixenv]
\matrix[mymatrix]  (m) 
{
1 & 0 & 0 & [3mm] \ldots & 0 & 1 & 0 & [3mm] \ldots & 0 & 0 & 1 & [3mm] \ldots & 0 & 0 & 0    \\
};
  \mymatrixbracetop{1}{3}{time 0}
  \mymatrixbracetop{5}{7}{time 4.5}
  \mymatrixbracetop{9}{11}{time 7.5}
  \mymatrixbracetop{13}{15}{time 12}
  
  \mymatrixbracebottom{1}{3}{run mixer}
  \mymatrixbracebottom{5}{7}{mixer}
  \mymatrixbracebottom{9}{11}{purificator}
  \mymatrixbracebottom{13}{15}{no process to run}

    \end{tikzpicture}
\caption{Instruction vector example for 12H time horizon.}
\label{fig:instruction_vector}
\end{figure*}

The objective of the case study in Section \ref{sec:case_study} is to maximise the amount of product produced in a given time horizon. As such, the simulation takes as input a schedule (herewith referred to as the instruction vector) instructing it at which times the processes should be started. The indices of the instruction vector represent the 0.5H time steps from 0, e.g. $0, 0.5, 1, 1.5, 2, \dots$. Each element in the instruction vector is a binary vector of size 3. Each index of this binary vector represents one of the processes (0 for purification, 1 for reaction, and 2 for mixing). Consider an instruction vector \textbf{x} = [[1,0,0], [0,0,0]], then at time 0, the instructions are [1,0,0] indicating that mixing should occur at time 0, while reaction and purification should not. At time 0.5, none of the processes would occur; however, since mixing started at time 0, it will run for its allotted time. If a process starts to run at time $t$, it will not be allowed to run again until its' scheduled time is complete. For a time horizon $n$, the instruction vector would have a length of $(n \times 2) \times 3$. For example, a 12H time horizon would have an instruction vector of length 72. An example instruction vector for the 12H time horizon is show in \ref{fig:instruction_vector}.

The simulation in this instance is not necessarily cheaper than the objective function, however for more complex scenarios (e.g. the multi-purpose batch processing problem \cite{MWoolway2018, MWoolway2019, MIerapetritou1998}, the simulation would be cheaper and the purpose of this study is to investigate if a cheap simulation can approximate an objective function well enough for future work. The simulation is an approximation because instruction vectors don't always produce the optimal objective value, only the optimal instruction vectors do.

Adjusting the systems' parameters creates a variation of the motivating example (called the primary example) with a stricter bottleneck (i.e. increasing the complexity of the optimisation problem). Reducing the storage capacities of State 2 and State 3 in Table \ref{tab:batch_processing} from 100 to 50 allows for more stringent bottleneck behaviour.

\subsection{Optimising the schedule of the multi-purposed batch-processing problem}

The batch-processing problem is a constrained maximisation optimisation problem with the number of variables for a given time horizon equal to the length of the instruction vector for that time horizon. The objective function to maximise is the output of the simulation described in Section \ref{sec:simulation} (i.e. the amount of material produced in a given time horizon). The simulation is coded in such a way that if a solution is infeasible (i.e. a process is scheduled to run but violates the system's constraints), then it accepts the solution but won't run any processes that violate the constraints of the system. Since the instruction vectors are binary, the lower and upper limits for the variables are 0 and 1, respectively.

The initial population is a randomly generated instruction vector with at most half the values set to 1 (with the rest being 0). Crossover and mutation are applied to generate the offspring population.

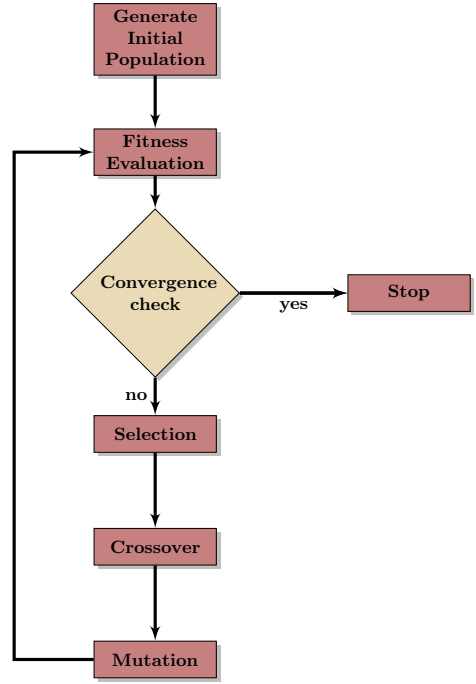
\begin{figure*}[htb!]
\centering
\scalebox{0.75}{
\tikzstyle{startstop} = [rectangle, draw, minimum height=2em, text centered, fill=darkred!50, text width=6em, drop shadow]

\tikzstyle{process} = [rectangle, draw, text width=6em, minimum height=2em, text centered, fill=darkred!50, drop shadow]
\tikzstyle{decision} = [diamond, draw, text width=6em, minimum height=2em, text centered, draw=black,text width=6em, fill=darkgoldenrod!30, drop shadow]
\tikzstyle{line} = [draw, -latex']

\begin{tikzpicture}[node distance=2cm]
\node (start) [startstop] {\textbf{Generate Initial Population}};

\node (fiteval) [process, below of=start] {\textbf{Fitness Evaluation}};

\node (convchk) [decision, below of=fiteval, yshift=-0.5cm] {\textbf{Convergence check}};

\node (selection) [process, below of=convchk, yshift=-0.5cm] {\textbf{Selection}};
\node (crossover) [process, below of=selection] {\textbf{Crossover}};
\node (mutation) [process, below of=crossover] {\textbf{Mutation}};

\node (stop) [startstop, right of=convchk, xshift=2.5cm] {\textbf{Stop}};

\draw [line, ultra thick] (start) -- (fiteval);

\draw [line, ultra thick] (fiteval) -- (convchk);

\draw [line, ultra thick] (convchk) -- node[anchor=north] {\textbf{yes}} (stop);
\draw [line, ultra thick] (convchk) -- node[anchor=east] {\textbf{no}} (selection);

\draw [line, ultra thick] (selection) -- (crossover);
\draw [line, ultra thick] (crossover) -- (mutation);
\draw [line, ultra thick] (mutation) -- +(-2.5,0) |- (fiteval);
\draw [line, ultra thick] (convchk) -- (stop);

\end{tikzpicture}
}

\caption{GA flowchart.}
\label{fig:ga_flowchart}
\end{figure*}

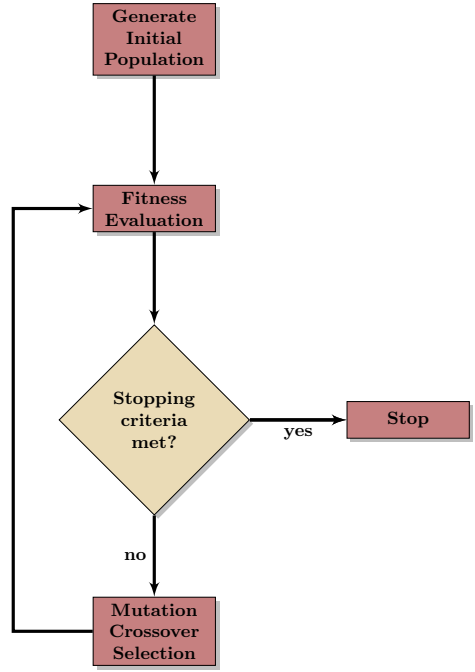
\begin{figure*}[htb!]
\centering
\scalebox{0.75}{
\tikzstyle{startstop} = [rectangle, draw, minimum height=2em, text centered, fill=darkred!50, text width=6em, drop shadow]

\tikzstyle{process} = [rectangle, draw, text width=6em, minimum height=2em, text centered, fill=darkred!50, drop shadow]
\tikzstyle{decision} = [diamond, draw, text width=6em, minimum height=2em, text centered, draw=black,text width=6em, fill=darkgoldenrod!30, drop shadow]
\tikzstyle{line} = [draw, -latex']

\begin{tikzpicture}[node distance=3cm, scale=1]
\node (start) [startstop] {\textbf{Generate Initial Population}};

\node (fiteval) [process, below of=start] {\textbf{Fitness Evaluation}};

\node (convchk) [decision, below of=fiteval, yshift=-.75cm] {\textbf{Stopping criteria met?}};

\node (prcss) [process, below of=convchk, yshift=-.75cm] {\textbf{Mutation \\ Crossover \\ Selection}};

\node (stop) [startstop, right of=convchk, xshift=1.5cm] {\textbf{Stop}};

\draw [line, ultra thick] (start) -> (fiteval);

\draw [line, ultra thick] (fiteval) -> (convchk);

\draw [line, ultra thick] (convchk) -> node[anchor=north] {\textbf{yes}} (stop);
\draw [line, ultra thick] (convchk) -> node[anchor=east] {\textbf{no}} (prcss);

\draw [line, ultra thick] (prcss) -- +(-2.5,0) |- (fiteval);
\draw [line, ultra thick] (convchk) -> (stop);

\end{tikzpicture}
}
\label{fig:de_flowchart}
\caption{DE flowchart.}

\end{figure*}

The evolutionary algorithms studied in this paper are GA and DE. The algorithmic processes for GA and DE are shown in Figures \ref{fig:ga_flowchart} and \ref{fig:de_flowchart}, respectively. The PSAF framework, described in Section \ref{sec:psaf}, provides surrogate assistance to the evolutionary algorithms. Various quality indicators allow for comparing the quality of the solutions generated by the evolutionary algorithms and their surrogate-assisted counterparts. The quality indicators employed in this research are Success Rate (SR), Average Evaluations to a Solution (AESR) and Average Generations to a Solution (AGSR). SR measures the percentage of trials in which the best value returned by the algorithm was within a specific percentage of the optimal objective value. AESR measures the average number of evaluations required to reach a certain percentage of the optimal objective value. AGSR is the average number of generations to reach a certain percentage of the optimal objective value. For the primary example, 95\% and 99.5\% are chosen. For the variation of the problem, 90\% and 95\% are chosen. For robustness, each experiment is run 30 times, and the average value of each quality indicator is presented.

The various algorithms use the same parameters: the initial population size is 30, the number of generations is 20, and the number of offspring is 10. For both the primary example and the variation, the parameters for PSAF are; $\alpha$ is 5, $\beta$ is 5, and the number of infills is 10. It should be noted that the parameters have not been tuned, which could improve results. For the 168H time horizon, the number of generations is reduced to 15. Table \ref{tab:neww} provides the parameters used in the GA and DE algorithms for all the time horizons.

\section{Results}

\begin{table*}[htb!]
    \caption{Primary Example Quality Indicators.}
    \begin{center}
        \resizebox{\columnwidth}{!}{%
        \begin{tabular}{lcrrrrrrr}
            \toprule
            \textbf{Algorithm}  & 
            \textbf{\makecell[cb]{Time\\Horizon}} & 
            \textbf{\makecell[rb]{Objective\\Value}} & 
            \textbf{\makecell[cb]{SR\\@95}}     & 
            \textbf{\makecell[cb]{SR\\@99.5}} & 
            \textbf{\makecell[cb]{AESR\\@95}}&
            \textbf{\makecell[cb]{AESR\\@99.5}} & 
            \textbf{\makecell[cb]{AGSR\\@95}}& 
            \textbf{\makecell[cb]{AGSR\\@99.5}} \\[3pt]\\
            \toprule
            \multirow{7}{*}[-.1cm]{GA}      & 12H          & 100             & 100       & 100         & 30          & 30           & 1  & 1        \\[3pt]
                                                                                           & 24H          & 350             & 100       & 100         & 30          & 30           & 1 & 1         \\[3pt]
                                                                                           & 36H          & 625             & 100.00       & 100.00         & 44.67       & 56.67  & 2.83      & 3.83       \\[3pt]
                                                                                           & 48H          & 900             & 100.00       & 73.33       & 34.00          & 107.72   & 1.47    & 8.46      \\[3pt]
                                                                                           & 60H          & 1150            & 100.00       & 60.00          & 30.00          & 142.22  & 1.00      & 12.22      \\[3pt]
                                                                                           & 72H          & 1425            & 100.00       & 36.67       & 34.00          & 140.00    & 1.40      & 12.00         \\[3pt]
                                                                                           & 168H         & 3550            & 100       & 0         & 30        & 0   & 1      & 0         \\[3pt]
           \bottomrule
            \multirow{7}{*}[-.1cm]{PSAF-GA} & 12H          & 100             & $100.00^{=}$ & $100.00^{=}$   & $31.33^{-}$ & $31.33^{-}$  & $1.13^{-}$ & $1.13^{-}$ \\[3pt]
                                                                                           & 24H          & 350             & $100^{=}$ & $100^{=}$   & $30^{=}$    & $30^{=}$     & $1^{=}$ & $1^{=}$   \\[3pt]
                                                                                           & 36H          & 625             & $100.00^{=}$ & $100.00^{=}$   & $32^{+}$    & $48.67^{+}$  & $1.2^{+}$ & $2.87^{+}$ \\[3pt]
                                                                                           & 48H          & 900             & $100.00^{=}$ & $53.33^{-}$ & $40.67^{-}$ & $76.25^{+}$ & $2.07^{-}$ & $5.63^{+}$ \\[3pt]
                                                                                           & 60H          & 1150            & $100.00^{=}$ & $33.33^{-}$ & $36.67^{-}$ & $82.00^{+}$ & $1.67^{-}$    & $6.20^{+}$  \\[3pt]
                                                                                           & 72H          & 1425            & $100.00^{=}$ & $20.00^{-}$    & $31.33^{+}$ & $106.67^{+}$ & $1.13^{+}$ & $8.67^{+}$ \\[3pt]
                                                                                           & 168H         & 3550            & $100.00^{=}$         & $0.00^{=}$           & $34.44^{-}$           & $0.00^{=}$   & $1.44^{-}$        & $0.00^{=}$          \\[3pt]

            \toprule
            \multirow{7}{*}[-.1cm]{DE}  & 12H          & 100             & 100.00       & 100.00         & 30.33         & 30.33    & 1.03      & 1.03          \\[3pt]
            
                                                                                           & 24H          & 350             & 100       & 100         & 30         & 30          & 1  & 1        \\[3pt]
            
                                                                                           & 36H          & 625             & 100.00       & 73.33         & 30.00         & 30.00          & 1.00 & 1.00         \\[3pt]
            
                                                                                           & 48H          & 900             & 100.00       & 80.00         & 32.00         & 123.33    & 1.20      & 10.33          \\[3pt]
          
                                                                                           & 60H          & 1150            & 100.00       & 33.33         & 30.00         & 135.00    & 1.00      & 11.50          \\[3pt]
            
                                                                                           & 72H          & 1425            & 100.00       & 0.00         & 30.67         & 0.00       & 1.07   & 0 .00        \\[3pt]
            
                                                                                           & 168H         & 3550            & 100       & 0         & 30        & 0      & 1   & 0         \\[3pt]
            \toprule
            \multirow{7}{*}[-.1cm]{PSAF-DE} & 12H          & 100             & $100.00^{=}$ & $100.00^{=}$   & $30.67^{-}$ & $30.67^{-}$ & $1.07^{-}$ & $1.07^{-}$ \\[3pt]
            
                                                                                           & 24H          & 350             & $100.00^{=}$ & $100.00^{=}$   & $31.33^{=}$    & $31.33^{-}$     & $1.13^{-}$ & $1.13^{-}$    \\[3pt]
          
                                                                                           & 36H          & 625             & $100.00^{=}$ & $100.00^{+}$   & $30.00^{=}$    & $52.67^{-}$  & $1.00^{=}$ & $3.27^{-}$ \\[3pt]
            
                                                                                           & 48H          & 900             & $100.00^{=}$ & $100.00^{+}$ & $44.00^{-}$ & $83.33^{+}$ & $2.40^{-}$ & $6.33^{+}$ \\[3pt]
           
                                                                                           & 60H          & 1150            & $100.00^{=}$ & $26.67^{-}$ & $30.00^{=}$ & $132.50^{+}$    & $1.00^{=}$ & $11.25^{+}$  \\[3pt]
           
                                                                                           & 72H          & 1425            & $100.00^{=}$ & $6.67^{+}$    & $34.67^{-}$ & $140.00^{+}$ & $1.47^{-}$ & $12.00^{+}$ \\[3pt]
            
                                                                                           & 168H         & 3550            & $100^{=}$         & $100^{=}$           & $30^{=}$          & $30^{+}$   & $1^{=}$         & $1^{+}$        \\[3pt]

             \bottomrule                                                                              
            \multicolumn{4}{l}{$^{\mathrm{+}}$Surrogate improvement.}                                                                                                                           \\
            \multicolumn{4}{l}{$^{\mathrm{-}}$Surrogate deterioration.}                                                                                                                         \\
            \multicolumn{4}{l}{$^{\mathrm{=}}$Surrogate maintained.}
        \end{tabular}}
        \label{tab:primary_quality_indicators}
    \end{center}
\end{table*}

\begin{table*}[htb!]
    \caption{Variant Example Quality Indicators.}
    
    \begin{center}
        \resizebox{\columnwidth}{!}{%
        \begin{tabular}{ccccccccc}
            \toprule
            \textbf{Algorithm} & 
            \textbf{\makecell[cb]{Time\\Horizon}} & 
            \textbf{\makecell[rb]{Objective\\Value}} & 
            \textbf{\makecell[cb]{SR\\@90}}     & 
            \textbf{\makecell[cb]{SR\\@95}}  & 
            \textbf{\makecell[cb]{AESR\\@90}} &
            \textbf{\makecell[cb]{AESR\\@95}} & 
            \textbf{\makecell[cb]{AGSR\\@90}} & 
            \textbf{\makecell[cb]{AGSR\\@95}} \\[3pt]\\
            \toprule
            \multirow{7}{*}[-.1cm]{GA}      & 12H          & 100             & 100       & 100         & 30          & 30           & 1  & 1        \\[3pt]
                                                                                           & 24H          & 325             & 100.00 & 93.33       & 53.33         & 87.14                    & 1 & 4.21         \\[3pt]
                                                                                           & 36H          & 575             & 100.00       & 80.00         & 61.33       & 155.00  & 1.80      & 10.58       \\[3pt]
                                                                                           & 48H          & 800             & 100.00       & 33.33      & 56.67          & 742.00   & 1.33    & 25.40     \\[3pt]
                                                                                           & 60H          & 1000            & 100       & 100          & 30          & 50  & 1      & 3      \\[3pt]
                                                                                           & 72H          & 1250            & 100.00       & 93.33       & 32.00          & 104.28    & 1.20      & 8.43         \\[3pt]
                                                                                           & 168H         & 2825            & 100       & 100         & 30        & 30   & 1      & 1         \\[3pt]
            \toprule
            \multirow{7}{*}[-.1cm]{PSAF-GA}      & 12H          & 100             & $100^{=}$       & $100^{=}$         & $30^{=}$          & $30^{=}$           & $1^{=}$  & $1^{=}$        \\[3pt]
                                                                                           & 24H          & 325             & $100.00^{=}$       & $100.00^{+}$         & $30.00^{+}$          & $65.33^{+}$           & $1.00^{=}$ & $4.53^{-}$         \\[3pt]
                                                                                           & 36H          & 575             & $93.33^{-}$       & $40.00^{-}$         & $40.00^{+}$       & $58.33^{+}$  & $2.00^{-}$      & $3.83^{+}$       \\[3pt]
                                                                                           & 48H          & 800             & $100.00^{=}$       & $0.00^{-}$       & $35.33^{+}$          & $0.00^{-}$   & $1.53^{-}$    & $0.00^{-}$      \\[3pt]
                                                                                           & 60H          & 1000            & $100.00^{=}$       & $86.67^{-}$          & $30.00^{=}$          & $67.69^{-}$  & $1.00^{-}$     & $4.77^{-}$      \\[3pt]
                                                                                           & 72H          & 1250            & $100.00^{=}$       & $53.33^{-}$       & $30.00^{+}$          & $90.00^{+}$    & $1.00^{+}$      & $7.00^{+}$         \\[3pt]
                                                                                           & 168H         & 2825            & $100^{=}$       & $100^{=}$         & $30^{=}$        & $30^{=}$   & $1^{=}$      & $1^{=}$         \\[3pt]
            \toprule
            
            \multirow{2}{*}[-.1cm]{DE}      & 12H          & 100             & 100       & 100         & 30          & 30           & 1  & 1        \\[3pt]
                                                                                           & 24H          & 325             & 100.00       & 80.00         & 30.00          & 32.5           & 1 & 1.25         \\[3pt]
     
                                                                                           & 36H          & 575             & 100       & 80.00         & 38.00       & 78.33  & 1.80      & 5.83       \\[3pt]
                                                                                           & 48H          & 800             & 100.00       & 0.00       & 50.67          & 0.00   & 3.07    & 0.00      \\[3pt]
                                                                                           & 60H          & 1000            & 100       & 100          & 30          & 57.33  & 1.00      & 3.73      \\[3pt]
                                                                                           & 72H          & 1250            & 100.00       & 46.67       & 35.33          & 104.29    & 1.53      & 8.43         \\[3pt]
                                                                                           & 168H         & 2825            & 100       & 100         & 30        & 30   & 1      & 1         \\[3pt]
            \toprule
            \multirow{7}{*}[-.1cm]{PSAF-DE}      & 12H          & 100             & $100^{=}$       & $100^{=}$         & $30^{=}$          & $30^{=}$           & $1^{=}$  & $1^{=}$        \\[3pt]
                                                                                           & 24H          & 325             & $100.00^{=}$       & $100.00^{=}$         & $30.00^{=}$          & $52.00^{-}$           & $1.00^{=}$ & $3.20^{-}$         \\[3pt]
                                                                                           & 36H          & 575             & $100.00^{=}$       & $40.00^{-}$         & $48.00^{-}$       & $53.33^{+}$  & $2.8^{-}$      & $3.33^{+}$       \\[3pt]
                                                                                           & 48H          & 800             & $100.0^{=}$       & $0.0^{=}$       & $32.0^{+}$          & $0.0^{=}$   & $1.2^{+}$    & $0.0^{=}$      \\[3pt]
                                                                                           & 60H          & 1000            & $100.00^{=}$       & $100.00^{=}$          & $30.00^{=}$          & $40.67^{+}$  & $1.00^{+}$      & $2.07^{+}$      \\[3pt]
                                                                                           & 72H          & 1250            & $100.00^{=}$       & $73.33^{+}$       & $31.33^{+}$          & $124.55^{-}$    & $1.13^{+}$      & $10.45^{-}$         \\[3pt]
                                                                                           & 168H         & 2825            & $100^{=}$       & $100^{=}$         & $30^{=}$        & $30^{=}$   & $1^{=}$      & $1^{=}$         \\[3pt]
            
            \bottomrule

            \multicolumn{4}{l}{$^{\mathrm{+}}$Surrogate improvement.}                                                                                                                           \\
            \multicolumn{4}{l}{$^{\mathrm{-}}$Surrogate deterioration.}                                                                                                                         \\
            \multicolumn{4}{l}{$^{\mathrm{=}}$Surrogate maintained.}
        \end{tabular}}
        \label{tab:variant_quality_indicators}
    \end{center}
\end{table*}

\begin{table*}[htb!]
    \caption{Experimental Setup (Primary and Variation Examples).}
    \begin{center}
    \resizebox{\columnwidth}{!}{%
        \begin{tabular}{ccccc}
            \toprule
            \textbf{Time Horizon}  & \textbf{Algorithm} & \textbf{Generations} & \textbf{Population Size} & \textbf{Offspring}  \\[3pt]
            \toprule
            \multirow{2}{*}[-.1cm]{12H}        & GA               & 20               & 30           & 10              \\[3pt]
            & DE         & 20                & 30             & 10                         \\[3pt]
            
            \midrule
            \multirow{2}{*}[-.1cm]{{\centering 24H}}        & GA               & 20               & 30           & 10             \\[3pt]
            & DE         & 20                & 30             & 10                         \\[3pt]
            
            \midrule
            \multirow{2}{*}[-.1cm]{{\centering 36H}}        & GA               & 20               & 30           & 10              \\[3pt]
            & DE         & 20                & 30             & 10                        \\[3pt]

            \midrule
            \multirow{2}{*}[-.1cm]{{\centering 48H}}        & GA               & 20               & 30           & 10             \\[3pt]
            & DE         & 20                & 30             & 10                        \\[3pt]

            \midrule
            \multirow{2}{*}[-.1cm]{{\centering 60H}}        & GA               & 20               & 30           & 10             \\[3pt]
            & DE         & 20                & 30             & 10                        \\[3pt]

            \midrule
            \multirow{2}{*}[-.1cm]{{\centering 72H}}        & GA               & 20               & 30           & 10             \\[3pt]
            & DE         & 20                & 30             & 10                         \\[3pt]

            \midrule
            \multirow{2}{*}[-.1cm]{{\centering 168H}}        & GA               & 15               & 30           & 10              \\[3pt]
            & DE         & 15                & 30             & 10                         \\[3pt]
            
            \bottomrule
        \end{tabular}
        \label{tab:neww}
       }
    \end{center}
\end{table*}

Tables~\ref{tab:primary_quality_indicators} and \ref{tab:variant_quality_indicators} show the quality indicators for the primary and variant examples, respectively, for each time horizon. The SR@95 and SR@99.5 values show the percentage of trials wherein the highest objective value was within 95\% and 99.5\% of the optimal objective value, respectively. The AESR@95 and AESR@99.5 values show the average number of evaluations required to reach 95\% and 99.5\% of the optimal objective value, respectively. The AGSR@95 and AGSR@99.5 values show the average number of generations required to reach 95\% and 99.5\% of the optimal objective value, respectively. Studying the surrogate-assisted results, we expect to see the SR values increase (i.e. the success rates should increase). We also expect the AESR and AGSR values to decrease (i.e. it should take the surrogate-assisted algorithms fewer evaluations and generations to get close to the optimal objective value).

Table~\ref{tab:primary_quality_indicators} shows that PSAF-GA provides general improvement for the AESR@99.5 and AGSR@99.5 indicators. The SR@99.5 shows that PSAF-GA had much lower success rates than the GA. The AESR@95 shows mixed results wherein there is an improvement for the 36H, 72H and 168H time horizons, whereas there is deterioration and maintenance for the others. The AGSR@95 also shows an improvement for some time horizons and deterioration or maintenance for others. Interestingly, under the specified parameters, the GA and PSAF-GA could not reach 99.5\% of the optimal objective value for the 168H time horizon.

\begin{figure*}[htb!]
    \centering
    \resizebox{.95\textwidth}{!}{
	\begin{tikzpicture}
	\begin{axis}[
	xlabel = Generation,
	ylabel = Objective Value,
	axis background/.style={fill=gray!7, draw=gray},
                width=\textwidth,
                height=.9*\axisdefaultheight,
	legend pos=south east
	]
	\addplot[darkred, ultra thick, no marks, dotted] table [x=gens, y=OV, col sep=comma] {Plot_data/168_GA.csv};
	\addplot[moonstoneblue, ultra thick, dashed] table [x=gens, y=OV, col sep=comma] {Plot_data/168_DE.csv};
	\addplot[darkgoldenrod, ultra thick, dashdotdotted] table [x=gens, y=OV, col sep=comma] {Plot_data/168_PSAF-GA.csv};
	\addplot[darkcyan, ultra thick, no marks] table [x=gens, y=OV, col sep=comma] {Plot_data/168_PSAF-DE.csv};
	\legend{GA, DE, PSAF-GA, PSAF-DE};
	\end{axis}
	\end{tikzpicture}
	}
	\caption{(Primary Example) 168H Generations vs Objective Value.}
    \label{fig:generations_ov}
\end{figure*}
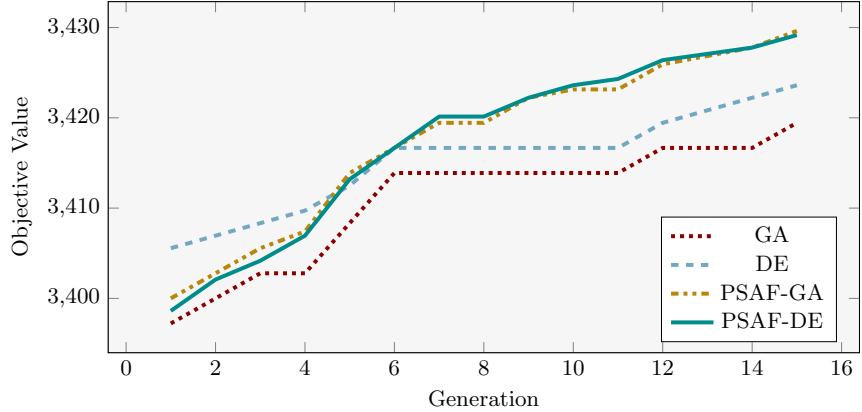

Comparing the DE and PSAF-DE shows a general improvement in the SR@99.5 metric. However, Table \ref{tab:primary_quality_indicators} shows mixed results in the other metrics where the surrogate-assisted PSAF-DE showed improvement for some time horizons and deterioration or maintenance for others. DE could not reach 99.5\% of the optimal objective value for the 72H and 168H time horizons, whereas PSAF-DE could. Figure \ref{fig:generations_ov} highlights that for the 168H time horizon, PSAF-DE and PSAF-GA can get closer to the optimal objective value quicker than the baseline GA and DE algorithms (i.e. it tends towards the optima in fewer generations). This result highlights the value of enhancing evolutionary algorithms through surrogate assistance.

Studying the results for the variation of the problem, Table \ref{tab:variant_quality_indicators} shows that in some cases, PSAF-GA provided an improvement over the baseline GA, whereas, in others, there was either maintenance or deterioration. Studying the results of PSAF-DE and the baseline DE paints a similar picture, wherein some of the test cases show an improvement by PSAF-DE, although that improvement is not visible in all test cases. Figure \ref{fig:generations_ov} shows that for the 168H time horizon in the primary example, PSAF-DE and PSAF-GA can converge towards the optimal objective value in fewer generations than DE and GA, respectively.
These results showcase the potential for using a surrogate-assisted framework but also highlight the importance of tuning the surrogates' hyper-parameters for achieving consistent, improved performance. Further, the results presented here warrant the exploration of using PSAF for solving the more general multi-purpose batch processing problem which is a multi-objective optimisation problem.

\section{Conclusion}

This paper studied the application of the evolutionary algorithms GA, DE, and their surrogate-assisted counterparts PSAF-GA and PSAF-DE to the batch-processing problem \cite{MIerapetritou1998, MWoolway2018, MWoolway2019}. 

The task is to maximise the product produced by the simulation for a given time horizon. The solutions generated by the evolutionary algorithms represent a schedule for the simulation to follow. This paper also considers a problem variation in which the systems' parameters are modified to enforce stricter bottleneck behaviour, thereby increasing the complexity of the optimisation task.

Table \ref{tab:primary_quality_indicators} shows the potential of a surrogate-assisted framework like PSAF for solving optimisation problems. In some cases, in the primary example, the surrogate improved over the baseline algorithm or maintained the quality of solutions (e.g. PSAF-DE showed improvement on AESR@99.5 and AGSR@99.5). In contrast, the surrogate deteriorated over the baseline algorithm in some cases (e.g. PSAF-DE showed deterioration on AESR@90 and AGSR@90). Table \ref{tab:variant_quality_indicators} shows similar behaviour in the variant of the variant example. Figure \ref{fig:generations_ov} shows that for the 168H time horizon in the primary example, PSAF-GA and PSAF-DE converge in fewer generations than GA and DE. These results show that it is feasible to build a simulation of the problem which can be used as the objective function in the optimisation problem.

Finally, these findings show that surrogate-assisted frameworks have the potential to improve baseline evolutionary algorithms. In future work, tuning the parameters of the PSAF framework, $\alpha$ and $\beta$, could provide further improvement by the surrogate. This paper considered the single-objective batch-processing problem, to explore the effectiveness of using a surrogate-assisted framework like PSAF to solve optimisation problems using a simulation as the objective function. The results presented here warrant the exploration of using PSAF for solving the more general multi-purpose batch processing problem which is a multi-objective optimisation problem.

\bibliographystyle{splncs04nat}
\bibliography{ref}

\end{document}